%% file: main.tex
\PassOptionsToPackage{warn}{textcomp}
\documentclass[sigconf]{acmart}

\usepackage[utf8]{inputenc}

\usepackage{graphicx}

\usepackage{caption}
\usepackage{subcaption}
\usepackage{float}
\usepackage{booktabs}

\usepackage{amsmath}
\usepackage{xcolor}

\usepackage{hyperref}
\usepackage{cleveref}

\usepackage{todonotes}      

\DeclareMathOperator{\E}{\mathbb{E}}
\DeclareMathOperator{\cov}{cov}
\DeclareMathOperator{\softmax}{\operatorname{Softmax}}
\DeclareMathOperator*{\argmax}{argmax}

\DeclareMathOperator{\exposure}{{Exp}}

\DeclareMathOperator{\utility}{\Delta}

\newcommand{\myvec}[1]{\boldsymbol{#1}}

\newcommand{\ranking}{\sigma}
\newcommand{\relevance}{\mathrm{rel}}

\newcommand{\merit}{\mathrm{M}}
\newcommand{\model}{FULTR}
\newcommand{\smodel}{\model\space}
\newcommand{\queryset}{\boldsymbol{Q}}

\copyrightyear{2021}
\acmYear{2021}
\setcopyright{acmcopyright}
\acmConference[SIGIR '21] {Proceedings of the 44th International ACM SIGIR Conference on Research and Development in Information Retrieval}{July 11--15, 2021}{Virtual Event, Canada.}
\acmBooktitle{Proceedings of the 44th International ACM SIGIR Conference on Research and Development in Information Retrieval (SIGIR '21), July 11--15, 2021, Virtual Event, Canada}
\acmPrice{15.00}
\acmISBN{978-1-4503-8037-9/21/07}
\acmDOI{10.1145/3404835.3462953}

\begin{document}
\fancyhead{}

\title{Policy-Gradient Training of Fair and Unbiased Ranking Functions}

\author{Himank Yadav}
\authornote{Equal contribution}
\email{himankyadav1@gmail.com}
\affiliation{%
  \institution{Cornell University}
  \city{Ithaca}
  \state{NY}
  \country{USA}
}

\author{Zhengxiao Du}
\authornotemark[1]
\email{zx-du20@mails.tsinghua.edu.cn}
\affiliation{%
  \institution{Tsinghua University}
  \city{Beijing}
  \country{China}
}

\author{Thorsten Joachims}
\email{tj@cornell.edu}
\affiliation{%
  \institution{Cornell University}
  \city{Ithaca}
  \state{NY}
  \country{USA}
}

\emergencystretch 3em%
\begin{abstract}
While implicit feedback (e.g., clicks, dwell times, etc.) is an abundant and attractive source of data for learning to rank, it can produce unfair ranking policies for both exogenous and endogenous reasons. Exogenous reasons typically manifest themselves as biases in the training data, which then get reflected in the learned ranking policy and often lead to rich-get-richer dynamics. Moreover, even after the correction of such biases, reasons endogenous to the design of the learning algorithm can still lead to ranking policies that do not allocate exposure among items in a fair way. To address both exogenous and endogenous sources of unfairness, we present the first learning-to-rank approach that addresses both presentation bias and merit-based fairness of exposure simultaneously. Specifically, we define a class of amortized fairness-of-exposure constraints that can be chosen based on the needs of an application, and we show how these fairness criteria can be enforced despite the selection biases in implicit feedback data. The key result is an efficient and flexible policy-gradient algorithm, called \model, which is the first to enable the use of counterfactual estimators for both utility estimation and fairness constraints. Beyond the theoretical justification of the framework, we show empirically that the proposed algorithm can learn accurate and fair ranking policies from biased and noisy feedback.\footnote{The code and datasets are available at \url{https://github.com/him229/fultr}.}

\end{abstract}



\maketitle

\input{introduction.tex}
\input{related_work.tex}
\input{preliminary.tex}
\input{fairness.tex}
\input{training.tex}

\input{experiment.tex}
\section{Conclusion}
We presented a framework \model~ for learning accurate and fair ranking policies from biased feedback. Specifically, we introduced fairness-of-exposure constraints that can allocate amortized exposure to groups of items based on their amortized relevance. Furthermore, we proposed counterfactual estimators of the corresponding utility and disparity measures that remove the effects of position bias and click noise. For efficient training, we derived a policy gradient algorithm that directly optimizes both utility and fairness. Finally, we presented extensive empirical evidence that \model\ can effectively learn ranking policies under fairness constraints despite biased and noisy feedback. Given that \model~ estimates relevance based on position bias across multiple queries, one existing limitation of the framework is it's inability to deal with individual-level fairness due to a lack of sufficient data. In the future, it would be interesting to explore ways to scale \model~ to incorporate individual fairness.

\begin{acks}
This research was supported in part by NSF Awards IIS-1901168
and IIS-2008139. All content represents the opinion of
the authors, which is not necessarily shared or endorsed by their
respective employers and/or sponsors.
\end{acks}

\bibliographystyle{ACM-Reference-Format}
\bibliography{reference.bib}


\end{document}

%% file: introduction.tex
\section{Introduction}

Implicit feedback from user behavior (e.g., clicks, dwell times, purchases, scroll patterns)~\cite{joachims2005accurately_implicit, rendle2009bpr_implicit} 
is an attractive source of training data for learning-to-rank (LTR), since it gives all users a voice in what the system learns. But does the participatory nature of implicit feedback automatically ensure that the learned ranking policies are fair and lead to desirable effects on the overall health of the ranking system? We argue that both endogenous and exogenous factors can lead to undesirable outcomes as follows.

{\em Endogenous} factors are design choices built into the allocation policies of the ranking system. Specifically, deciding which ranking to present to a user implies an explicit choice of how much exposure each item receives -- where higher-ranked items receive more exposure and thus more opportunity~\cite{craswell2008experimental, dupret2008user}. 
Conventional ranking algorithms~\cite{ProbRankPrinciple} make this exposure allocation in a haphazard way, 
which can lead to unfairness and undesirable dynamics~\cite{BiegaGW18:EquityAttention, FairnessExposure:SinghJ18, PolicyFairness:abs-1902-04056,beutel2019fairness,zehlike2017fa, celis2017ranking,GeyikAK19:LinkedIn,yang2017measuring}. 
To address this problem, many have noted that there should be an explicit link between the amount of exposure an item receives, and the amount of merit (e.g. relevance) the item has. Specifically, merit-based exposure allocation~\cite{BiegaGW18:EquityAttention, FairnessExposure:SinghJ18, PolicyFairness:abs-1902-04056,beutel2019fairness} can ensure that items of similar merit (e.g. relevance) also have similar outcomes (e.g. exposure). The particular choice of allocation policy can provide various fairness guarantees and provide a mechanism for steering the system dynamics (e.g. avoiding superstar economics~\cite{Mehrotra/etal/18}).

While merit-based exposure allocation addresses an endogenous factor of unfairness, it is important to recognize that an LTR algorithm can still be unfair due to {\em exogenous} biases in the training data. In particular, position bias~\cite{JoachimsGPHRG07:Implicit} makes the use of implicit feedback for LTR challenging. Especially for positive-only feedback like clicks or purchases, highly-ranked items receive more positive feedback due to the increased attention they receive, which further skews the ranking system and affects future rankings. The result can be a dynamic amplification of position bias, leading to phenomena like rich-get-richer~\cite{adamic2000power_rich, PropSVMRank:JoachimsSS17, JoachimsGPHRG07:Implicit, salganik2006experimental_rich}. 
This means the even in the presence of merit-based fairness constraints like~\cite{FairnessExposure:SinghJ18, PolicyFairness:abs-1902-04056}, naive and biased estimates of merit from implicit feedback signals can still lead to inequities.
We, therefore, need to design fair LTR algorithms that ensure unbiased learning despite the biases in the data.

In this paper, we simultaneously address endogenous and exogenous causes of unfairness and present the first LTR algorithm -- called \model~(Fair Unbiased Learning-to-Rank) -- for unbiased learning from logged implicit feedback that can enforce merit-based exposure-allocation constraints. 
We provide a theoretical analysis to show that a naive combination of conventional merit-based fairness constraints and counterfactual relevance estimators cannot guarantee unbiased and fair ranking policies.
To overcome this problem, we define a new type of allocation constraint for groups of items and derive counterfactual estimators~\cite{SelectionBias:WangBMN16, PropSVMRank:JoachimsSS17} that provably correct the position bias. To make the proposed framework operational and practical, we show how to search the space of fairness-constrained ranking policies via a novel policy-gradient approach that is unbiased. Note that \model\ is trained end-to-end, and thus can consider fairness during learning, eliminating failure cases of two-step approaches that rely on post-processing an unfair ranking to make it fair~\cite{BiegaGW18:EquityAttention,FairnessExposure:SinghJ18,Morik/etal/20a}. 
 
Beyond the theoretical justification, we also provide an extensive empirical evaluation on real-world datasets. We find that \model\ can effectively optimize utility and fairness over a range of settings even when trained with biased and noisy feedback.
\vspace{-1em}

\subsection{A Motivating Example}

Consider an online movie streaming service which presents a ranked list of movies for any incoming query. Further, consider the case where the movies were produced by only two studios. Studio $G_1$ produced 30 movies that all have relevance $0.7$ (i.e. there is a 70\% probability that the user will want to watch each movie). Studio $G_2$ also produced 30 movies, but they all have slightly lower relevance of $0.69$. Ranking purely based on relevance to the user would place movies from $G_1$ at spots $1-30$ and movies from $G_2$ at spots $31-60$, leading to much higher exposure and revenue for movies from $G_1$ than movies from $G_2$. This can be considered unfair since movies of almost equal merit (i.e. relevance) receive disproportionately worse outcomes (i.e. streaming revenue). 

A first thought for addressing this shortcoming may be to enforce various forms of demographic parity between groups of items in the ranking~\cite{zehlike2017fa, celis2017ranking,GeyikAK19:LinkedIn,yang2017measuring}. However, allocating exposure without consideration of merit has other undesirable effects. Assume that there is a third, incompetent, studio $G_3$ that produced another $30$ movies of low relevance $0.1$. Clearly, these movies do not deserve as much exposure as those from $G_1$ and $G_2$.
We therefore follow an approach that explicitly links exposure to the merit (i.e. relevance) of a group. Specifically, merit-based exposure allocation~\cite{BiegaGW18:EquityAttention, FairnessExposure:SinghJ18, PolicyFairness:abs-1902-04056,beutel2019fairness} can ensure that groups of similar merit (e.g. relevance) have similar outcomes (e.g. exposure). 

Enforcing merit-based fairness of exposure is a factor endogenous to the ranking system, meaning that it is fully under the control of the system designer. However, it relies on accurate estimates of merit, which can be affected by exogenous factors. Considers, for example, a situation where movies from $G_1$ have consistently been ranked higher than those from $G_2$ in the past. If we now naively count the number of clicks as an estimate of merit, this estimate will be biased by how humans browse ranked lists \cite{JoachimsGPHRG07:Implicit}. In particular, the movies from $G_2$ are likely to get substantially fewer clicks than those from $G_1$ not because they are substantially worse, but because they were less likely to be discovered by the users. In this way, even an endogenously fair ranking system can become unfair as its past actions create biased estimates that amplify past inequities and that can more generally lead to undesirable system dynamics \cite{Mehrotra/etal/18,PropSVMRank:JoachimsSS17}.

Our approach is the first that can simultaneously control both sources of unfairness when learning a ranking policy, as we will detail in the following.

%% file: related_work.tex
\section{Related Work}


There have been numerous approaches to defining fairness in different areas of machine learning, including online learning~\cite{gillen2018online,patil2019achieving_smab}, classification~\cite{agarwal2018_class,narasimhan2018learning_class}, regression~\cite{rezaei2019fair_reg,berk2017convex_reg}. We focus on fairness in the relatively under-explored domain of LTR, which has only recently caught attention despite its substantial implications in a broad range of real-world applications. To structure the discussion, we follow the distinction of endogenous and exogenous sources of unfairness introduced above. 

Concerning endogenous fairness, several methods have followed the concept of demographic parity which enforces a proportional allocation of exposure between groups without considering merit. This is achieved by either reducing the difference in occurrences of different groups on a subset of the rankings~\cite{yang2017measuring} or by placing a limit on the number of items from each group in the top-k positions~\cite{zehlike2017fa, celis2017ranking,GeyikAK19:LinkedIn}. In contrast, merit-based fairness of exposure \cite{FairnessExposure:SinghJ18,BiegaGW18:EquityAttention} allocates exposure to groups based on their merit rather than purely based on group size. However, these methods assume knowledge of the true relevance labels for all items and do not involve learning, while in practice, true relevance labels are not available. and one needs to use other learning methods to impute relevance labels at prediction time~\cite{Morik/etal/20a}. This leads to a two-step process, where the first regression step is unaware of fairness considerations and fairness is only introduced during post-processing. Instead, our method is trained end-to-end in one step, such that it can, for example, discount features that lead to unfair relevance estimates~\cite{PolicyFairness:abs-1902-04056}.

More recently, \citet{zehlike2018reducing} proposed an LTR method that incorporates an exposure-based fairness regularizer into the ranking loss, but it only considers the exposure at the top-1 position. The limitation was removed by \citet{PolicyFairness:abs-1902-04056}, who propose an end-to-end LTR method that optimizes both utility and fairness constraints for the full ranking. However, both methods assume that expert-labeled relevances for all items is available during training. This is a weaker assumption than the one made in \cite{FairnessExposure:SinghJ18,BiegaGW18:EquityAttention}, but it still does not apply to real-world settings where implicit feedback is often used for training.


Our work focuses on exposure-based fairness, but another approach to defining fairness in rankings is through pairwise comparisons~\cite{beutel2019fairness, narasimhan2019pairwise}, without considering exposure as the key criterion. Such comparisons between items count all swaps in the ranking equally and do not reflect that swapping items at the top has a stronger effect on exposure than doing so at lower positions.

Switching to exogenous sources of unfairness, most work on traditional LTR algorithms~\cite{LambdaRank:BurgesRL06, CaoQLTL07:ListNet, SVMRank:Joachims02, BurgesSRLDHH05:RankNet} assumes that unbiased relevance judgments by experts are available. However, the field of information retrieval has long been conscious of the biases inherent to implicit feedback and its effect on the ability to rank well~\cite{SVMRank:Joachims02,JoachimsGPHRG07:Implicit,craswell2008experimental, dupret2008user, guo2009efficient}. In particular, various studies have demonstrated the presence of position bias, where the quantity and quality of the feedback depend on the rank at which an item is presented. 

One approach to modeling and removing position bias is generative click modeling~\cite{chapelle2009dynamic_click,borisov2016neural_click,chuklin2015click}. Click models typically treat relevance as latent variables, and perform inference by maximizing log-likelihood of clicks. These inferred relevances can then be used as a substitute for expert labels in LTR.
Unfortunately, most click models suffer from the limitation of requiring large amounts of repeat impressions for individual query-item pairs, which makes them inapplicable to tail queries.

Recent and more direct approaches to dealing with position bias are counterfactual learning methods \cite{PropSVMRank:JoachimsSS17, SelectionBias:WangBMN16}. These methods use techniques from causal inference like inverse propensity score (IPS) weighting~\cite{PropensityScore} and do not require repeated queries or latent-variable inference. Instead, they directly optimize over a debiased utility objective that incorporates click data in a principled fashion.
Additional algorithms~\cite{ai2018unbiased_joint,hu2018novel_joint} that jointly estimate the propensities and optimize the performance have been proposed. Unlike our proposed \model~framework, existing counterfactual learning methods do not control for endogenous unfairness. Moreover, we find that naively combining existing counterfactual learning techniques with endogenous fairness constraints fail to control for unfairness.


The only existing method that aims to address both endogenous and exogenous sources of unfairness in rankings is \cite{Morik/etal/20a}. 
However, it is the two-step method that learns relevance estimation with unbiased regression objectives first. The regression learning is unaware of fairness constraints and may yield unfairness in relevance estimation. We show empirically that our end-to-end \model~ can achieve better fairness-utility tradeoff than the two-step method. Additionally, it requires interactive experimental control in the form of an online algorithm, whereas \model~can reuse logged implicit feedback data from past interactions for learning. 
To the best of our knowledge, no other existing method addresses fairness in rankings while directly learning from logged implicit feedback. 

%% file: preliminary.tex
\section{Learning Fair Ranking Policies from Implicit Feedback}
\label{sec:fairness} \label{sec:preliminary}

In the following, we derive the first learning algorithm for enforcing merit-based fairness of exposure~\cite{FairnessExposure:SinghJ18} with logged implicit feedback. A key component is a new amortized fairness criterion, and its corresponding disparity measure, for which we show that unbiased counterfactual estimators exist. We analyze the connection between this disparity measure and those in \cite{FairnessExposure:SinghJ18,BiegaGW18:EquityAttention}. Furthermore, we quantify the effect of click noise on the disparity measure and propose a noise-corrected estimator of disparity.

The derivation of the algorithm is based on the standard definition of Learning to rank (LTR), which is the problem of learning a ranking policy $\pi$ for queries $q$ sampled from a distribution $P(q)$. Each query $q$ has a set of candidate items $\myvec{d}^q$ that need to be ranked.
Items can represent a wide variety of things depending on the application --- web-pages in a search engine, jobs on a job board, or movies in a streaming service. 
Each item is associated with a feature vector $x_{q,d}$ that describes the match between item $d$ and query $q$. 

In the so-called full-information setting, it is assumed that for a training query $q$ the relevances $\relevance_{q,d}$ of all items $d\in \myvec{d}^q$ are known. Instead, implicit feedback only reveals the relevances of part of the candidates. We denote this partial feedback as $c_q$ and call this the partial-information setting~\cite{PropSVMRank:JoachimsSS17}. 

Unlike most existing works, we consider stochastic ranking policies, where $\pi(\ranking|q)$ is a distribution over the rankings $\ranking$ of the candidates. As will become clear later, stochastic ranking policies have the advantage of providing more fine-grained control of exposure and enabling gradient-based optimization. Note that deterministic ranking policies are a special case of stochastic ranking policies, where all probability mass lies on a single ranking.

Like in conventional LTR algorithms~\cite{LambdaRank:BurgesRL06, CaoQLTL07:ListNet, SVMRank:Joachims02, BurgesSRLDHH05:RankNet}, our objective is to learn a policy $\pi$ from policy space $\Pi$ with high expected utility 
\begin{equation*}
    U(\pi)=\E_{q\sim P(q)}[U(\pi|q)]=\E_{q\sim P(q)}\E_{\ranking\sim \pi(\ranking|q)}\left[\utility(\ranking,\relevance_q)\right],
\end{equation*}
where $\utility$ can be any ranking metric (e.g. DCG~\cite{DCG:JarvelinK02}).  

Utility optimization for the users, however, does not by itself ensure fairness to the items~\cite{FairnessExposure:SinghJ18}. We, therefore, include an additional constraint that addresses endogenous unfairness by enforcing a merit-based allocation of exposure. For simplicity, consider the case of two groups $G_i$ and $G_j$. A group is simply a collection of items by any criterion, such as gender, price, brand, etc. To ensure that exposure is allocated fairly between $G_i$ and $G_j$, we measure unfairness via a disparity measure $D_{ij}(\pi)$ that will be defined in \Cref{subsec:amortized_fairness}. A perfectly fair ranking policy has zero disparity. More generally, we want to restrict $[D_{ij}(\pi)]^2$ to be at most some threshold $\delta$ to get our disparity to be as close to zero as possible. We define our objective by combining the fairness constraint with the conventional utility optimization. 
\begin{equation}
	\pi^* = \argmax_{\pi\in\Pi} U(\pi)\;
\text{s.t.}\; [D_{ij}(\pi)]^2 \leq \delta.
\end{equation}
Directly optimizing this objective is not possible, since neither the query distribution $P(q)$ nor the relevances $\relevance_q$ are known. We therefore take the following Empirical Risk Minimization (ERM) approach, where both $U(\pi)$ and $D_{ij}(\pi)$ are replaced by estimates $\widehat{U}(\pi|\queryset)$ and $\widehat{D}_{ij}(\pi|\queryset)$ based on a partial-info training set $\queryset$.
\begin{equation}
	\label{eqn:partial_obj}
	\widehat{\pi} = \argmax_{\pi\in\Pi} \widehat{U}(\pi|\queryset)\;
	\text{s.t.}\; \big[\widehat{D}_{ij}(\pi|\queryset)\big]^2 \leq \delta.
\end{equation}
This leads to three key challenges which we address as follows. First, in \Cref{subsec:unbiased_utility}, we show how to design $\widehat{U}(\pi|\queryset)$ to get unbiased utility estimates even if we do not have access to true relevance labels $\relevance_q$ but only have access to biased implicit feedback $c_q$. Second, in \Cref{subsec:amortized_fairness}, we show how to define estimators of the fairness disparity $\widehat{D}_{ij}(\pi|\queryset)$ that ensure merit-based exposure allocation while also being unbiased even with only the implicit feedback $c_q$. Finally, in \Cref{sec:training}, we show how to efficiently solve the resulting training problem from Equation~\eqref{eqn:partial_obj}.

\subsection{Unbiased Utility Estimator}
\label{subsec:unbiased_utility}
We consider the class of additive ranking measures that can be expressed as 
\begin{equation*}
    \utility(\ranking, \relevance_q)= \sum_{d\in \myvec{d}^q}\left[{f\left(\ranking(d)\right)\cdot\relevance_{q,d}}\right],
\end{equation*}
where $\ranking(d)$ denotes the rank of item $d$ in ranking $\ranking$ and $f()$ can be any weighting function. For the DCG metric, $f(\ranking(d))={1}/{\log_2\left(1+\ranking(d)\right)}$, while for the Average Rank metric, $f(\ranking(d))=-\ranking(d)$. For simplicity, we assume binary relevances $\relevance_{q,d}\in\{0,1\}$.

In the partial-information setting, the relevances $\relevance_q$ are not known and we can only observe $c_q$. What can we infer from $c_q$ about $\relevance_q$? We defer the discussion of noise to Section~\ref{subsec:click_noise} and consider the standard model that users click if and only if the result is relevant (i.e. $\relevance_{q,d}=1$) and examined by the user. 
We introduce the binary random variable $o_{q,d}$ indicating whether the item $d$ is examined by the user. Based on this, we can model the "propensity" $p(o_{q,d}=1|\ranking_q)$, of observing $\relevance_{q,d}$ given that ranking $\ranking_q$ was presented when the implicit feedback was logged. The propensity can be modelled in various ways \cite{PropSVMRank:JoachimsSS17, FangAJ19:ContextPBM}. Most common is the position-based examination model, where the propensity depends on only the position $\ranking_q(d)$ of $d$ in the ranking. This means $p(o_{q,d}=1|\ranking_q)=v_{\ranking_q(d)}$, where $v_k$ denotes the examination probability at position $k$. These position biases $v_k$ can be estimated with swap experiments~\cite{PropSVMRank:JoachimsSS17} or intervention harvesting~\cite{EsitimationNoIntrusive:AgarwalZWLNJ19}. With knowledge of the propensity, we can use Inverse Propensity Score (IPS) weighting to arrive at the following estimator for $\utility$~\cite{PropSVMRank:JoachimsSS17}
\begin{equation}
	\widehat{\utility}(\ranking, c_q)=\sum_{d:c_{q,d}=1}\frac{f\left(\ranking(d)\right)}{p(o_{q,d}=1|\ranking_q)}.
\end{equation}
The estimator is unbiased if all propensities are bounded away from zero~\cite{PropSVMRank:JoachimsSS17}. Furthermore, we can define an estimator of the utility as
\begin{equation}
	\widehat{U}(\pi|\queryset)=\frac{1}{|\queryset|}\sum_{q\in\queryset}\widehat{U}(\pi|q)=\frac{1}{|\queryset|}\sum_{q\in\queryset}\E_{\ranking\sim\pi(\ranking|q)}[\widehat{\utility}(\ranking,c_q)].
\end{equation}	
It is easy to verify that this estimator inherits unbiasedness from $\widehat{\utility}$. It is also consistent under the same condition on the propensities, thus implying that $\widehat{U}(\pi|\queryset)$ will converge to $U(\pi)$ as $|\queryset|$ increases.





%% file: fairness.tex

\subsection{Unbiased Fairness Constraints}
\label{subsec:amortized_fairness}

To remedy that the utility-maximizing ranking policy can be unfair even if the utility $U(\pi)$ is perfectly known for all $\pi \in \Pi$ \cite{FairnessExposure:SinghJ18}, we enforce additional criteria of how exposure is allocated based on merit (i.e. relevance). In our training problem from \Cref{eqn:partial_obj}, this is implemented through the disparity measure $\widehat{D}_{ij}(\pi|\queryset)$ in the constraint, which we now define formally. 

As a first step, we define the merit of an item $d_i$ as its relevance, i.e. $\merit_q(d)=\relevance_{q,d}\in\{0,1\}$. The exposure of an item $d$ in ranking $\ranking$ is defined as the probability of a user examining the item in the ranking. This is identical to the examination probability $p(o_{q,d}=1|\ranking)$ defined in \Cref{subsec:unbiased_utility}, but this model is now applied to all rankings instead of just the logged ranking $\ranking_q$. The exposure of $d$ under a stochastic ranking policy $\pi$ for a query $q$, denoted as $\exposure_q(d|\pi)$, is the expected exposure over all the possible rankings. Formally,
\begin{equation*}
    \exposure_q(d|\pi)=\E_{\ranking\sim\pi(\ranking|q)}\left[p(o_{q,d}=1|\ranking)\right]
\end{equation*}

Furthermore, the exposure of group $G_i$ is the aggregate of the exposure of the group members, $\exposure_q(G_i|\pi)=\sum_{d\in G_{i}^q}\exposure_q(d|\pi),$ where $G_i^q=G_i\cap \myvec{d}^q$.
Similarly, we define the merit of group $G_i$ for query $q$ as $\merit_q(G_i)=\sum_{d \in G_{i}^q} \merit_q(d)$. 

With these definitions in hand, we define our fairness disparity of policy $\pi$ as $D_{ij}(\pi) = \E_{q \sim P(q)}[D_{ij}(\pi|q)]$, where $D_{ij}(\pi|q)$ measures the disparate exposure of $G_i$ and $G_j$ based on their merit for query $q$ as
\begin{equation}
	\label{eqn:disparity_measure}
	D_{ij}(\pi|q) =\merit_q({G_j})\exposure_q(G_i|\pi) - \merit_q({G_i})\exposure_q(G_j|\pi).
\end{equation}
As discussed in more detail in the next section, this disparity measure formalizes the principle that proportionality between merit and exposure 
should hold for a fair ranking \cite{FairnessExposure:SinghJ18,BiegaGW18:EquityAttention}. While $\exposure_q(G_i|\pi)$ is an expectation that can be computed, we do have to estimate $D_{ij}(\pi)$ with respect to the unknown query distribution and the unknown merits. We therefore take the estimator of $D_{ij}(\pi)$ as
\begin{equation}
	\label{eqn:group_disparity_emp}
		\widehat{D}_{ij}(\pi|\queryset) = \frac{1}{|\queryset|}\sum_{q\in\queryset}\widehat{D}_{ij}(\pi|q).
\end{equation}
Furthermore, in the partial-information setting, we can get an unbiased estimate of $\merit_q(G_i)$ using the IPS estimator $\widehat{\merit}_q(G_i)=\sum_{d \in G_i^q}\frac{c_{q,d}}{p(o_{q,d}=1|\ranking_q)}$.
This means that $\merit_q(G_i)$ in \Cref{eqn:disparity_measure} is replaced with $\widehat{\merit}_q(G_i)$ to arrive at the following empirical disparity measure
\begin{equation}
	\label{eqn:group_disparity_estimator}
		\widehat{D}_{ij}(\pi|q) = \widehat{\merit}_q({G_j})\exposure_q(G_i|\pi) - \widehat{\merit}_q({G_i})\exposure_q(G_j|\pi).
\end{equation}
Note that $\widehat{D}_{ij}(\pi|q)$ is unbiased since $\widehat{\merit}_q(G_i)$ is unbiased and the exposure terms are non-random constants.
\begin{equation*}
	\begin{split}
		\E_{o_q}\!\!\left[\!\widehat{D}_{ij}(\pi|q)\!\right]\!=&\E_{o_q}\![\widehat{\merit}_q(\!G_j\!)\!]\exposure\!_q\!(G_i|\pi)-\E_{o_q}\![\widehat{\merit}_q(\!G_i\!)\!]\exposure\!_q\!(G_j|\pi)\\
		=&\:\merit_q(G_j)\exposure_q(G_i|\pi)-\merit_q(G_i)\exposure_q(G_j|\pi)
	\end{split}
\end{equation*}
Since $\widehat{D}_{ij}(\pi|q)$ is unbiased for each query $q$, the aggregate $\widehat{D}_{ij}(\pi|\queryset)$ is also unbiased for $D_{ij}(\pi)$, i.e. $\E_q\E_{o_q}[\widehat{D}_{ij}(\pi|\queryset)]=\E_q[{D}_{ij}(\pi|q)]=D_{ij}(\pi)$.
Furthermore, through the use of Hoeffding bounds, it is possible to show that $\widehat{D}_{ij}(\pi|\queryset)$ converges to the true disparity $D_{ij}(\pi)$ as $|\queryset|$ increases.

\subsection{Comparison with other Disparity Measures}
\label{subsec:disparity}

In this section, we further highlight the differences between our disparity measure from \Cref{eqn:disparity_measure} and various other disparity measures. 

The Disparate Treatment constraint from \cite{PolicyFairness:abs-1902-04056}
\begin{equation*}
   \forall q: \frac{\exposure_q(G_i|\pi)}{\merit_q({G_i})}=\frac{\exposure_q(G_j|\pi)}{\merit_q({G_j})},
\end{equation*}
expresses that for a given query, the exposure of each group should be proportional to its merit. However, evaluating this constraint requires knowledge of the full relevances which is infeasible to decipher from real-world implicit feedback. Directly replacing $\relevance_{q,d}$ with $c_{q,d}$ when evaluating $\merit_q(G_j)$ leads to biased estimation due to selection bias~\cite{PropSVMRank:JoachimsSS17}. To combat this, naively plugging in the IPS estimator $\widehat{\merit}_q(G_i)$ into the corresponding disparity measure
\begin{equation}
    \label{eqn:naive_disparity_estimator}
	\widehat{D}''_{ij}(\pi)= \frac{\exposure_q(G_i|\pi)}{\widehat{\merit}_q({G_i})}-\frac{\exposure_q(G_j|\pi)}{\widehat{\merit}_q({G_j})}
\end{equation}
cannot eliminate bias effectively, since generally $\E[{1}/{\widehat{\merit}_q(G_i)}]\neq{1}/\E[{\widehat{M}_q(G_i)}]$. For an unbiased estimate of ${1}/{\merit_q(G_i)}$ from $c_q$, we need to know the joint distribution of $o_{q,d}$ for all the items in $G_i^q$, which is difficult to model. Instead, the proposed disparity measure in \Cref{eqn:disparity_measure} has an unbiased estimator (\Cref{eqn:group_disparity_estimator}) to overcome the selection bias, which we find to provide substantial empirical benefits in \Cref{subsec:fairness-evaluation}.


Another proposed disparity measure using implicit feedback replaces query-specific exposure and merit with an amortized notion \cite{BiegaGW18:EquityAttention} over the query distribution. The disparity measure 
\begin{equation}
	\label{eqn:disp2}
	D'_{ij}(\pi)=\frac{\E_{q}[\exposure_q(G_i|\pi)]}{\E_{q}[\merit_q({G_i})]}-\frac{\E_{q}[\exposure_q(G_j|\pi)]}{\E_{q}[\merit_q({G_j})]}
\end{equation}
expresses that for all the queries, the amortized exposure of each group should be proportional to its amortized merit. We will show how $D'_{ij}(\pi)$
is related to our disparity measure $D_{ij}(\pi)$ from \Cref{eqn:disparity_measure} under two conditions. First, assume that the total exposure of both groups is a constant $\exposure_q(G_i)+\exposure_q(G_j)=C_E$. In practice, the top items receive most of the users' attention, so the total exposure for each query is relatively stable even if the size of the candidate set varies. Second, assume that the covariance of the exposure of group $G_j$ and the total number of relevant items in both groups, $\cov[\exposure(G_j|\pi),\merit_q(G_i)+\merit_q(G_j)]$, is zero. One sufficient condition for this assumption to be approximately satisfied is when the numbers of relevant items do not vary much between queries. Under these two assumptions, the disparity measures $D_{ij}(\pi)$ and $D'_{ij}(\pi)$ are equivalent up to a constant coefficient.
This can be shown by transforming the disparity measure $D'_{ij}(\pi)$ as follows
\begin{align*}
	\frac{D_{ij}(\pi)+\cov[\exposure_q(G_j|\pi), \merit_q(G_i)]-\cov[\exposure_q(G_i|\pi), \merit_q(G_j)]}{\E_{q}[\merit_q(G_i)]\E_{q}[\merit_q(G_j)]}.
\end{align*}
The difference between the covariance terms is zero under the two assumptions. Therefore, we have $D_{ij}(\pi)=\E_q[{\merit_q(G_i)}]\E_q[\merit_q(G_j)]D'_{ij}(\pi)$.

However, \cite{BiegaGW18:EquityAttention} requires the knowledge of true relevance labels and does not involve learning. Extending this approach to learn relevance labels can suffer from the failure case of a two-step approach that relies on post-processing an unfair relevance estimation to make it fair, since fairness constraints are not taking into account during the learning process. We can get a consistent estimator of $D'_{ij}(\pi)$ by using the IPS estimator of merits and replacing the expectations of exposure and merit with the average over the dataset
\begin{equation*}
	\widehat{D}'_{ij}(\pi|\queryset)=\frac{\sum_{q\in\queryset}\exposure_q(G_i|\pi)}{\sum_{q\in\queryset}\widehat{\merit}_q({G_i})}-\frac{\sum_{q\in\queryset}\exposure_q(G_j|\pi)}{\sum_{q\in\queryset}\widehat{\merit}_q({G_j})}.
\end{equation*}
As $N$ increases, $\widehat{D}'_{ij}(\pi|\queryset)$ will converge to $D'_{ij}(\pi)$. However, the estimator is biased, which means $\E[\widehat{D}'_{ij}(\pi|\queryset)]\neq D'_{ij}(\pi)$. This bias might increase the error of the estimator. We therefore conclude that our new disparity $D_{ij}(\pi)$ with its unbiased estimator $\widehat{D}_{ij}(\pi)$ in \Cref{eqn:group_disparity_estimator,eqn:group_disparity_emp} is preferable.

\subsection{Incorporating Click Noise}
\label{subsec:click_noise}
Up to now, we assumed that positive feedback $c_{q,d}=1$  reveals the relevance label $\relevance_{q,d}$ in a noise-free way -- precisely that $c_{q,d}=1$ if and only if $\relevance_{q,d}=1$ and $o_{q,d}=1$. In practice, the user might make mistakes when examining the relevances of items, and we now define the following noise model. With $1\ge\epsilon_+>\epsilon_-\ge 0$, we have
\begin{align*}
    P\left(c_{q,d}=1|\relevance_{q,d}=1, o_{q,d}=1\right) = \epsilon_+\\
    P\left(c_{q,d}=1|\relevance_{q,d}=0, o_{q,d}=1\right) = \epsilon_-
\end{align*}
If $\epsilon_+<1$, users might ignore relevant items even after examination. Otherwise if $\epsilon_->0$, users might give false positive feedback to irrelevant items after examination. Fortunately, the IPS estimator of utility is order-preserving for two policies $\pi_1$ and $\pi_2$ under click noise~\cite{PropSVMRank:JoachimsSS17}, namely
\begin{align*}
	&\mathbb{E}_q\Big[\widehat{U}(\pi_1|\queryset)\Big]>\E_q\Big[\widehat{U}(\pi_2|\queryset)\Big]\Leftrightarrow U(\pi_1)>U(\pi_2|q).
\end{align*}
Therefore, given enough data, the click noise does not affect the ability to find the ranking policy with optimal utility.

In contrast, the disparity estimator in \Cref{eqn:group_disparity_emp,eqn:group_disparity_estimator} is not order-preserving when $\epsilon_->0$: 
\begin{align*}
	\E_q\!\!\left[\widehat{D}_{ij}(\pi_1|\queryset)-\widehat{D}_{ij}(\pi_2|\queryset)\right]& \!=(\epsilon_+\!-\!\epsilon_-)\E_{q}\!\!\left[D_{ij}(\pi_1|\queryset)-D_{ij}(\pi_2|\queryset)\right]\\
	&\!\!\!\!\!\!\!\!\!\!\!\!\!\!\!+\epsilon_-\E_{q}\!\!\left[|G_j^q|\delta\exposure_q(G_i)-|G_i^q|\delta\exposure_q(G_j)\right],
\end{align*}
where $\delta\exposure_q(G_i)$ denotes ${\exposure_q(G_i|\pi_1)}-{\exposure_q(G_i|\pi_2)}$. However, if we know the level of negative noise $\epsilon_-$, we can correct the IPS estimator for group disparity as
\begin{equation}
	\label{eqn:noise-free}
	\begin{split}
		\widehat{D}_{ij}(\pi|q,\epsilon_-)=\;&{\widehat{\merit}_q(G_j)}{\exposure_q(G_i|\pi)}-{\widehat{\merit}_q(G_i)}{\exposure_q(G_j|\pi)}\\
	&-\epsilon_-\left(|G_j^q|{\exposure_q(G_i|\pi)}-|G_i^q|{\exposure_q(G_j|\pi)}\right).
	\end{split}
\end{equation}
We can prove that the corrected estimator is unbiased up to a constant factor that can be absorbed into the hyperparameter $\delta$ of \Cref{eqn:partial_obj}
\begin{equation*}
    \E_{c_q}\left[\widehat{D}_{ij}(\pi|q,\epsilon_-)\right]=(\epsilon_+-\epsilon_-)D_{ij}(\pi|q).
\end{equation*}.

To estimate the probability $\epsilon_-$ for false-positive noise, we can use a simple intervention similar to \cite{SelectionBias:WangBMN16, PropSVMRank:JoachimsSS17}. While identifying relevant items is difficult, identifying irrelevant items is typically easy (e.g., documents that contain none of the query terms). Therefore, we can insert an item $d$ that is known to be irrelevant at position $k$ of the ranking. The expected clickthrough rate for the item is $p(c_{q,d}=1|\relevance_{q,d}=0)=v_k\cdot \epsilon_-$.
This means that given the position bias and the clickthrough rate estimated from the intervention data, we can compute the noise level $\epsilon_-$. Note that this intervention only needs to be performed on a small set of users and only affects the quality of one position.

%% file: training.tex
\section{Policy-Gradient Algorithm for Fair LTR} \label{sec:training}
In the previous section, we defined a general framework for learning ranking policies from biased and noisy feedback under amortized fairness constraints. However, we still need an efficient algorithm to search the specific policy space for the solution of the training problem in \Cref{eqn:partial_obj}. To this effect, we first define a stochastic ranking policy space based on the Plackett-Luce ranking model as in \cite{PolicyFairness:abs-1902-04056} and then present a policy-gradient algorithm that optimizes the training objective.

\subsection{Plackett-Luce Ranking Model}
We define a stochastic ranking policy space $\Pi$ based on the Plackett-Luce model~\cite{Luce59, Plackett}. Specifically, each ranking policy $\pi \in \Pi$ is defined by a scoring function $h_\theta$, which can be any differentiable model with parameters $\theta$. $h_\theta$ takes the feature vectors $x_q$ of all the items for the current query $q$ as input and outputs a vector of scores $h_\theta(x_q)=(h_\theta(x_{q,d_1}),h_\theta(x_{q,d_2}),\cdots,h_\theta(x_{q,d_{n_q}}))$. Based on this score vector, the probability $\pi_\theta(r|q)$ of a ranking $r=<d_1, d_2,\cdots,d_{n_q}>$ is defined as the product of softmax distributions
\begin{equation}
	\pi_\theta(r|q)=\prod_{i=1}^{n_q}\frac{\exp(h_\theta(x_{q,d_i}))}{\sum_{j=i}^{n_q}\exp(h_\theta(x_{q,d_j})))}.
\end{equation}
To sample rankings from $\pi_\theta(r|q)$, we can sample items from the distribution $\softmax(h_\theta(x_q))$ without replacement and rank the items in the order in which they are drawn.

\subsection{Policy Gradient Training Algorithm} 
We use a Lagrange multiplier to solve the constrained optimization problem from \Cref{eqn:partial_obj}: $\hat{\pi} = \argmax_{\pi}\min_{\lambda\ge 0} \widehat{U}(\pi|\queryset) - \lambda \Big( \big[\widehat{D}_{ij}(\pi|\queryset)\big]^2 - \delta\Big)$. Instead of solving the minimization problem w.r.t. $\lambda$, we search a specific range of $\lambda\in\{\lambda_1,\cdots,\lambda_k\}$. For each $\lambda$, we need to solve
\begin{equation}
	\label{eqn:unconstrained_objective}
	\hat{\pi}_{\lambda} = \argmax_{\pi}\widehat{U}(\pi|\queryset) - \lambda\big[\widehat{D}_{ij}(\pi|\queryset)\big]^2.
\end{equation}
Afterwards, we can compute the corresponding $\delta_\lambda=\big[\widehat{D}_{ij}(\hat{\pi}_\lambda|\queryset)\big]^2$ for which the constraint holds and then pick the optimal $\hat{\pi}_\lambda$ that satisfies $\delta_\lambda\le\delta$ and provides maximal utility $\widehat{U}(\hat{\pi}_\lambda|\queryset)$.

It remains to find an efficient algorithm for solving the unconstrained optimization problem in Equation~\eqref{eqn:unconstrained_objective}. We use stochastic gradient descent (SGD) to iteratively update the parameters of the ranking policy. However, it is intractable to compute the expectations $\widehat{U}(\pi|q)$ and $\widehat{D}_{ij}(\pi|q)$ over the exponential space of rankings. Instead, we use the log-derivative trick of the REINFORCE algorithm~\cite{Reinforce:Williams92} to compute the gradient of $\widehat{U}$
\begin{equation}
	\label{eqn_utility_grad}
	\begin{split}
		\nabla_\theta \widehat{U}(\pi_\theta|\queryset) =& \frac{1}{|\queryset|}\nabla_\theta \sum_{q}  \E_{\ranking \sim \pi(\ranking|q)} \big[\widehat{\Delta}(\ranking, c_q)\big]\\
		=& \frac{1}{|\queryset|} \sum_{q}\E_{\ranking \sim \pi(\ranking|q)} [\nabla_\theta \log \pi_\theta(\ranking|q) \widehat{\Delta}(\ranking, c_q)].
	\end{split}
\end{equation}
The gradient of the squared fairness disparity, $\nabla_\theta \big[\widehat{D}_{ij}(\pi_\theta|\queryset)\big]^2$, can also be reformulated this way,
\begin{equation}
	\label{eqn:disparity_grad}
	\begin{split}
		\frac{2}{|\queryset|}\widehat{D}_{ij}(\pi_\theta|\queryset)\sum_{q\in\queryset}\E_{\ranking\sim\pi_\theta(\ranking|q)}\left[\nabla_\theta \log \pi_\theta(\ranking|q)\widehat{\text{diff}}_{ij}(\ranking|q)\right],
	\end{split}
\end{equation}
where $\widehat{\text{diff}}_{ij}(\ranking|q)$ denotes $\widehat{M}_q(G_j)\exposure_q(G_i|\ranking) - \widehat{M}_q(G_i)\exposure_q(G_j|\ranking)$. The expectations over rankings in \Cref{eqn_utility_grad,eqn:disparity_grad} are approximated via $S=32$ Monte-Carlo samples in our experiments.

To apply SGD, we need to compute $\widehat{D}_{ij}(\pi_{\theta_t}|\queryset)$ on the whole dataset at each step $t$, which is quite expensive. 
As a simple but practically effective solution, we use the running average $\frac{1}{n}\sum_{\tau=0}^{n-1}\widehat{D}_{ij}(\pi_{\theta_{t-\tau}}|q_{t-\tau})$ as an approximation of $\widehat{D}_{ij}(\pi_{\theta_t}|\queryset)$ in \Cref{eqn:disparity_grad}. This is biased since the disparities are computed on previous parameters, but it can reduce the variance of the unbiased estimator $\widehat{D}_{ij}(\pi_{\theta_t}|q_t')$ with $q_t'$ from the query set.

%% file: experiment.tex
\section{Empirical Evaluation}
We conduct various experiments on two real-world datasets from which we generate synthetic click data following the experiment setup in \cite{PropSVMRank:JoachimsSS17,agarwal2019addressing_trustbias}. This allows us to control the experimental conditions and test multiple data distributions to evaluate robustness.

\begin{figure*}[!ht]
	\centering
    \begin{subfigure}[b]{0.825\columnwidth}
         \centering
         \includegraphics[width=\textwidth]{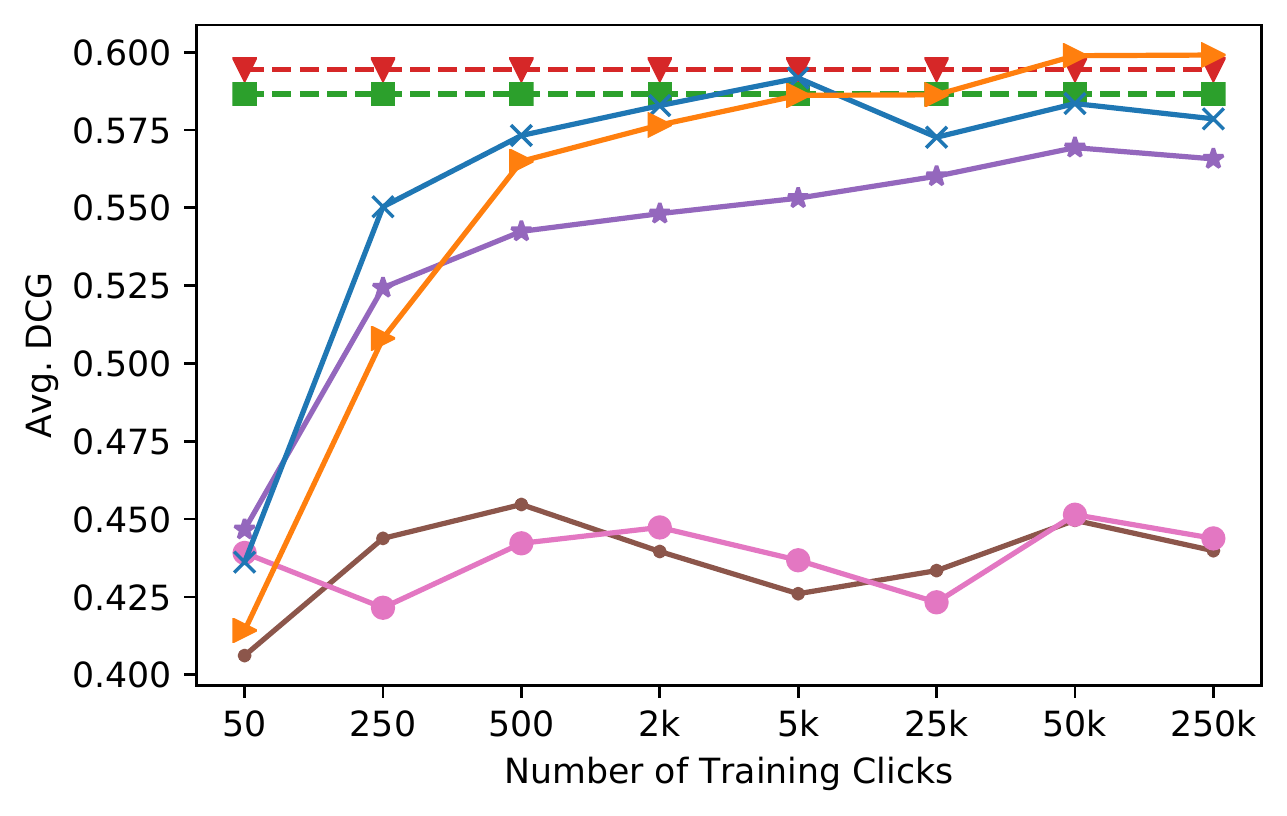}
         \caption{German Credit Dataset}
         \label{fig:dcg_german}
	\end{subfigure}
	\begin{subfigure}[b]{1.08\columnwidth}
		\centering
		\includegraphics[width=\textwidth]{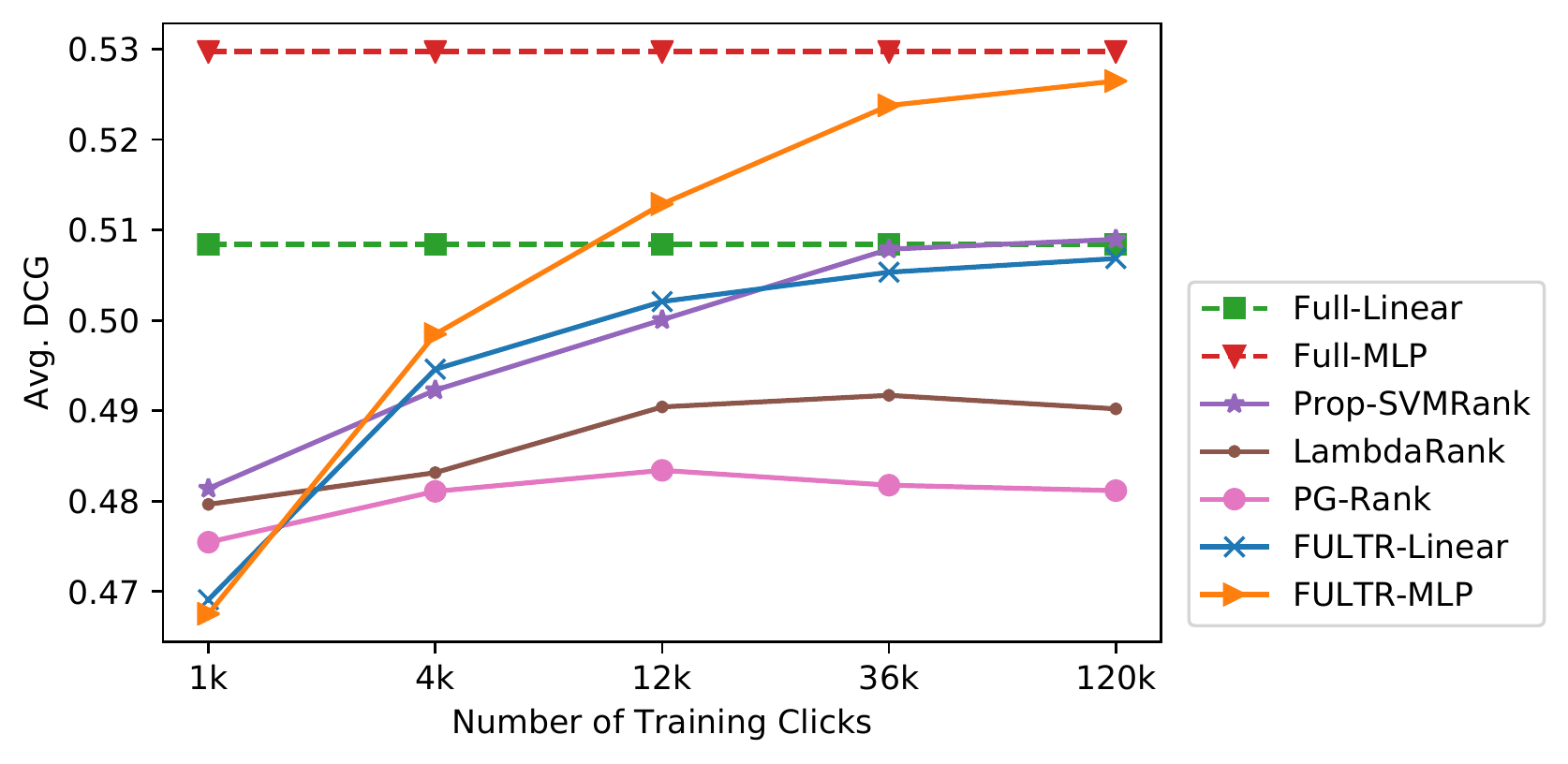}
		\caption{MSLR Dataset}
		\label{fig:dcg_mslr}
	   \end{subfigure}
	\caption{Ranking utility performance in terms of training clicks in the partial-info setting ($\eta=1,\epsilon_-=0.1$).}
	\label{fig:utility}
\end{figure*}
\subsection{Setup}
\subsubsection{Datasets}
We use the Microsoft Learning to Rank Dataset (Fold1), a large-scale LTR dataset~\cite{QinL13:LETOR}, and the German Credit Dataset, a credit scoring dataset often used to study algorithmic fairness~\cite{Dua:2019}.
The German Credit Dataset contains information about 1000 loan applicants, where each applicant is described by a set of attributes and labeled as creditworthy or non-creditworthy. We randomly split the applicants into train, validation, and test sets with ratio 1:1:1. To define groups, we use the binary feature A43 (indicating whether the purpose of the loan applicant is radio/television) as the group attribute. The ratio of applicants in two groups is around 8:2.

The Microsoft LTR Dataset contains a large number of queries from Bing with manually-judged relevance labels for retrieved webpages. We adopt the train, validation, and test split provided with the dataset. Since the dataset does not come with any designated groups, we select the QualityScore (feature id 133) as the group attribute and divide items into two groups with the 40th percentile on the attribute as the threshold. 

\subsubsection{Click Simulation}
We generate click data for training and validation from the full-information datasets following \cite{PropSVMRank:JoachimsSS17,agarwal2019addressing_trustbias}. We first train a Ranking SVM with 1 percent of the full-information training data as the logging policy. This logging policy is then used to generate the rankings for which click data is logged. The click data is generated by simulating the position-based examination model. The position bias decays with the presented rank $k$ of the item as $v_{k}=({1}/{k})^\eta$. When not stated otherwise, we use $\eta=1,\epsilon_+=1$, and $\epsilon_-=0$. Since the click signal is always binary and cannot reflect the relevance ratings, we binarize relevances in the same way as \cite{PropSVMRank:JoachimsSS17}, by assigning $\relevance_q(d)=1$ to all the items that were judged as 3 or 4 and $\relevance_q(d)=0$ to judgments 0, 1, and 2.

We preprocess both datasets into similar formats for LTR.  Following \cite{PolicyFairness:abs-1902-04056}, we build a query for the German Credit Dataset by sampling 20 individuals from the corresponding set with ratio 9:1 for non-creditworthy individuals to creditworthy individuals respectively. We sample 500 queries from train, validation, and test sets respectively. For MSLR, after the binarization step, the dataset becomes extremely sparse, with only about 2.5\% relevant items per query. To better compare different methods and amplify differences, for each query we sample 20 candidate items with 3 relevant items to build a new query.

The performance of the methods is reported on the full-information test set for which all relevance labels are known. Ranking utility and fairness are measured with Average DCG~\cite{DCG:JarvelinK02} and squared disparity $[D_{ij}(\pi)]^2$ respectively.

\subsubsection{Models and Hyperparameters} 
We train \smodel for two types of models: a linear model and a feed-forward network (one hidden layer with ReLU activation). We use the SGD optimizer for the linear model and the Adam optimizer for the neural network. The learning rate is 0.001. 
Following \cite{PolicyFairness:abs-1902-04056}, we subtract the average reward of the Monte-Carlo samples from the reward to act as a control variate for variance reduction~\cite{Reinforce:Williams92}. To encourage exploration and avoid premature convergence to suboptimal policies~\cite{MnihBMGLHSK16:A3C}, we add the entropy of the probability distribution $\softmax(h_\theta(x_q))$ to the objective, with a regularization coefficient $\gamma$. We initialize the coefficient as $\gamma=1.0$ and reduce it by a factor of 3 each time the validation metric has stopped improving. 
Finally, we add an L2 regularization term and cross-validate for the best regularization coefficient.

\subsection{Can \smodel learn accurate ranking policies from biased feedback?}
\label{subsec:utility-evaluation}
We begin our empirical evaluation by establishing that \model~without fairness constraints is competitive with conventional LTR methods when learning from partial feedback. Here, we ignore fairness and entirely focus on utility, just like in conventional LTR. For realism, we inject noise $\epsilon_-=0.1$. We compare \model~with the following LTR methods based on linear models or neural networks:
\begin{itemize}
    \item LambdaRank~\cite{LambdaRank:BurgesRL06}: This is a nonlinear ranking model based on neural networks to optimize DCG.
    \item Propensity SVM-Rank~\cite{PropSVMRank:JoachimsSS17}: This is an unbiased version of SVM-Rank~\cite{SVMRank:Joachims02} and utilizes the IPS estimator to correct position bias in implicit feedback.
    \item PG-Rank~\cite{PolicyFairness:abs-1902-04056}: This is a full-information policy-gradient method that is training the same model as \model-Linear.
\end{itemize}
Note that the full-information methods LambdaRank and PG-Rank ignore the position bias and treat click signals naively as full relevances. We also add a linear model and a neural network which get to see all true relevance labels as two skylines, called Full-Linear and Full-MLP. They represent the maximum performance we could hope to achieve with \model, which has access to only strictly less informative click feedback.

In \Cref{fig:utility}, we show the test-set performance of the conventional LTR methods compared to \model. Since the optimal policy is always deterministic, following \cite{PolicyFairness:abs-1902-04056}, we use the highest probability ranking for each query when evaluating \model. With increasing amounts of click data on the x-axis, \model~ with the neural network and the linear model approaches the skyline performance of Full-MLP and Full-Linear respectively on both datasets. The performance of \model-Linear is similar to Propensity SVM-Rank. Baseline methods that naively ignore position bias, namely LambdaRank and PG-Rank, cannot make effective use of the increasing amount of click data, and their learning curves are quite flat. Their performance is well below \model~given sufficient click data. 

Overall, we conclude that \model~is an LTR algorithm that achieves ranking performance that is at least competitive with existing baselines when learning linear or neural models from partial-information feedback. This implies that \model~is a promising contender for use in practical applications, and it thus makes sense to further investigate how far it can enforce fairness considerations.
\begin{figure*}[t]
	\centering
	\begin{subfigure}[b]{0.83375\columnwidth}
		\includegraphics[width=\textwidth]{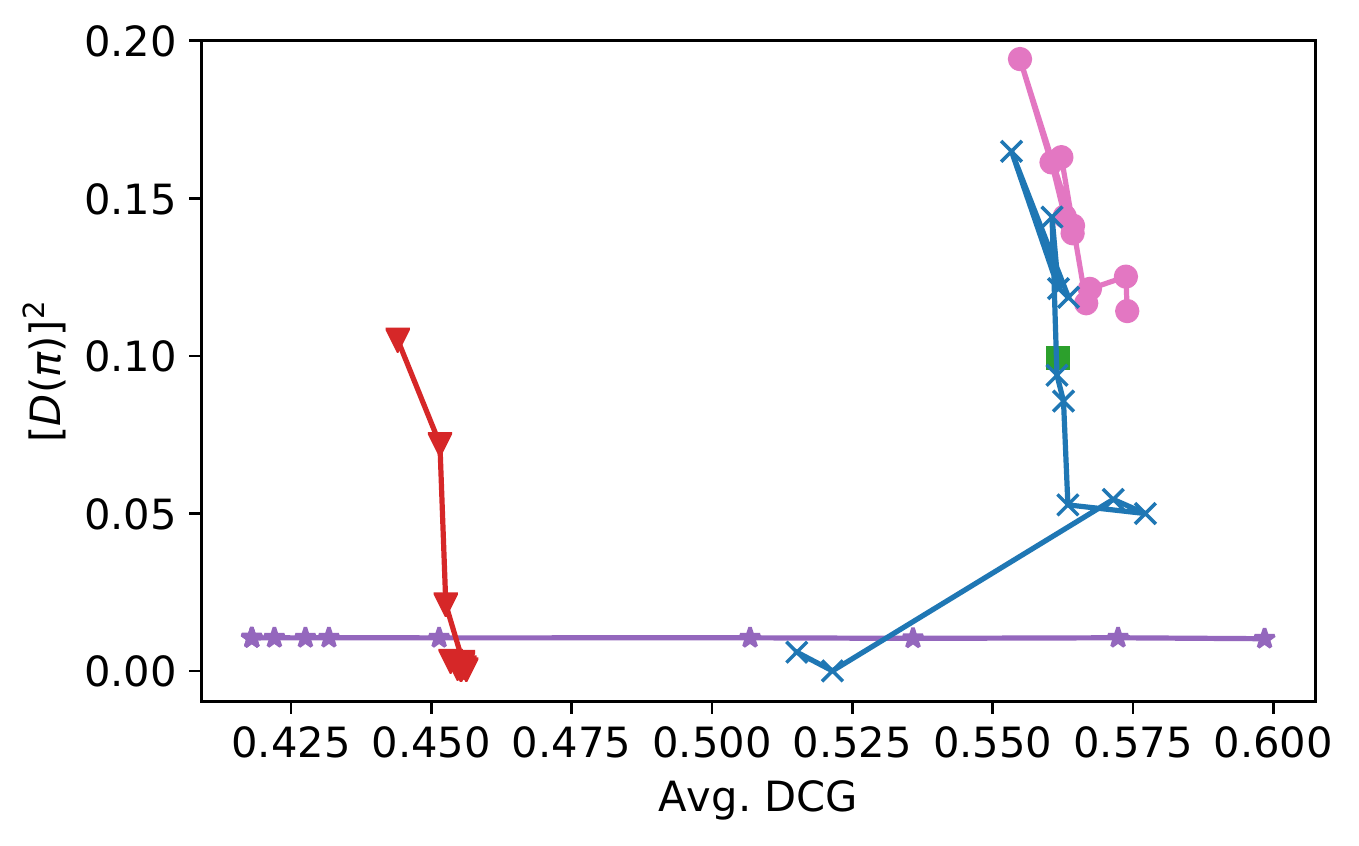}
		\caption{German Credit Dataset ($N=5K$)}
	\end{subfigure}
	\begin{subfigure}[b]{1.12\columnwidth}
		\includegraphics[width=\textwidth]{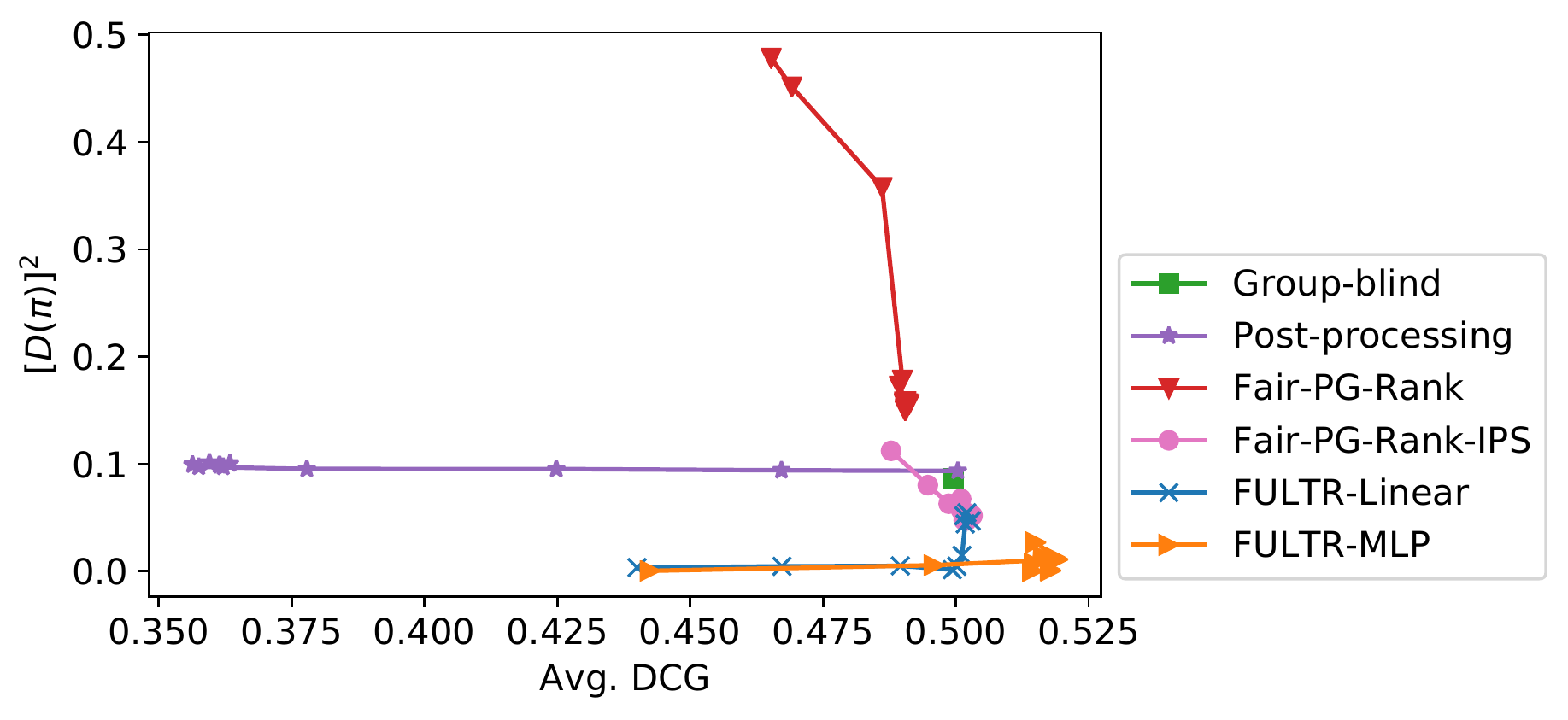}
		\caption{MSLR Dataset ($N=12K$)}
	\end{subfigure}
	\caption{DCG-Disparity curves for \model~ against baselines in the partial-info setting ($\eta=1,\epsilon_-=0$).}
	\label{fig:weight_tradeoff}
\end{figure*}

\subsection{Can \smodel learn fair ranking policies from biased feedback?}
\label{subsec:fairness-evaluation}
Next, we investigate \model's ability to enforce merit-based fairness of exposure. We compare with the following methods as baselines:
\begin{itemize}
	\item Group-blind: A variant of \model~without any fairness constraints, but with the group attribute masked (i.e., fairness through unawareness).
	\item Fair-PG-Rank~\cite{PolicyFairness:abs-1902-04056}: PG-Rank with a fairness constraint. 
	\item Fair-PG-Rank-IPS: Fair-PG-Rank ~\cite{PolicyFairness:abs-1902-04056}, naively using the IPS estimator as in \Cref{eqn:naive_disparity_estimator}.
	\item Equity of Attention \cite{BiegaGW18:EquityAttention}: Post-processing method that optimizes amortized group disparity. We train a regression model $r(x_{q,d})$ to predict item relevances using the position-bias-corrected objective for relevance estimation in \cite{Morik/etal/20a}.
\end{itemize}

The resulting utility/disparity curves on the test set are shown in \Cref{fig:weight_tradeoff}. First, note that group-blind training does not lead to merit-based fairness of exposure. Second, \model~ is able to effectively trade-off utility and fairness on both datasets. As $\lambda$ increases, the disparity goes down with an associated drop in DCG as expected. Increasing $\lambda$ further drives disparity close to zero. The neural network version, FULTR-MLP, achieves better utility-fairness trade-off than the linear version on MSLR dataset. The German Credit dataset was not large enough to train a reliable MLP model.

Third, Fair-PG-Rank cannot effectively reduce amortized disparity. On both datasets, as we vary the coefficient $\lambda$, utility drops while disparity increases. We conjecture that this is due to optimizing fairness per query instead of amortized fairness, and the propagation of bias in clicks to the merit-based disparity measure. The naive combination of Fair-PG-Rank and IPS weighting cannot reduce the disparity, since the disparity estimator is still biased.

Fourth, the post-processing method cannot reduce amortized disparity on the MSLR dataset but can approach zero disparity on German Credit dataset. To further analyze the reason, we ran the post-processing method with ground-truth relevances. This infeasible approach achieves disparities close to zero on MSLR, which demonstrates that the problem lies in the regression model. To investigate further, we compute the accumulated regression errors for both groups. We observe that on MSLR, the regression model underestimates the relevance of $G_i$ but estimates the relevance of $G_j$ precisely (error -14.05\% vs -0.09\%), which means that the regression model is already unfair despite its bias-corrected training objective. On the German Credit dataset, the regression model underestimates the relevance of two groups more uniformly (error -15.86\% vs -19.47\%). We conclude that the failure of the post-processing on MSLR is due to its two-step nature, where the regression model is trained oblivious to fairness.
\begin{figure}[!ht]
	\includegraphics[width=0.9\columnwidth]{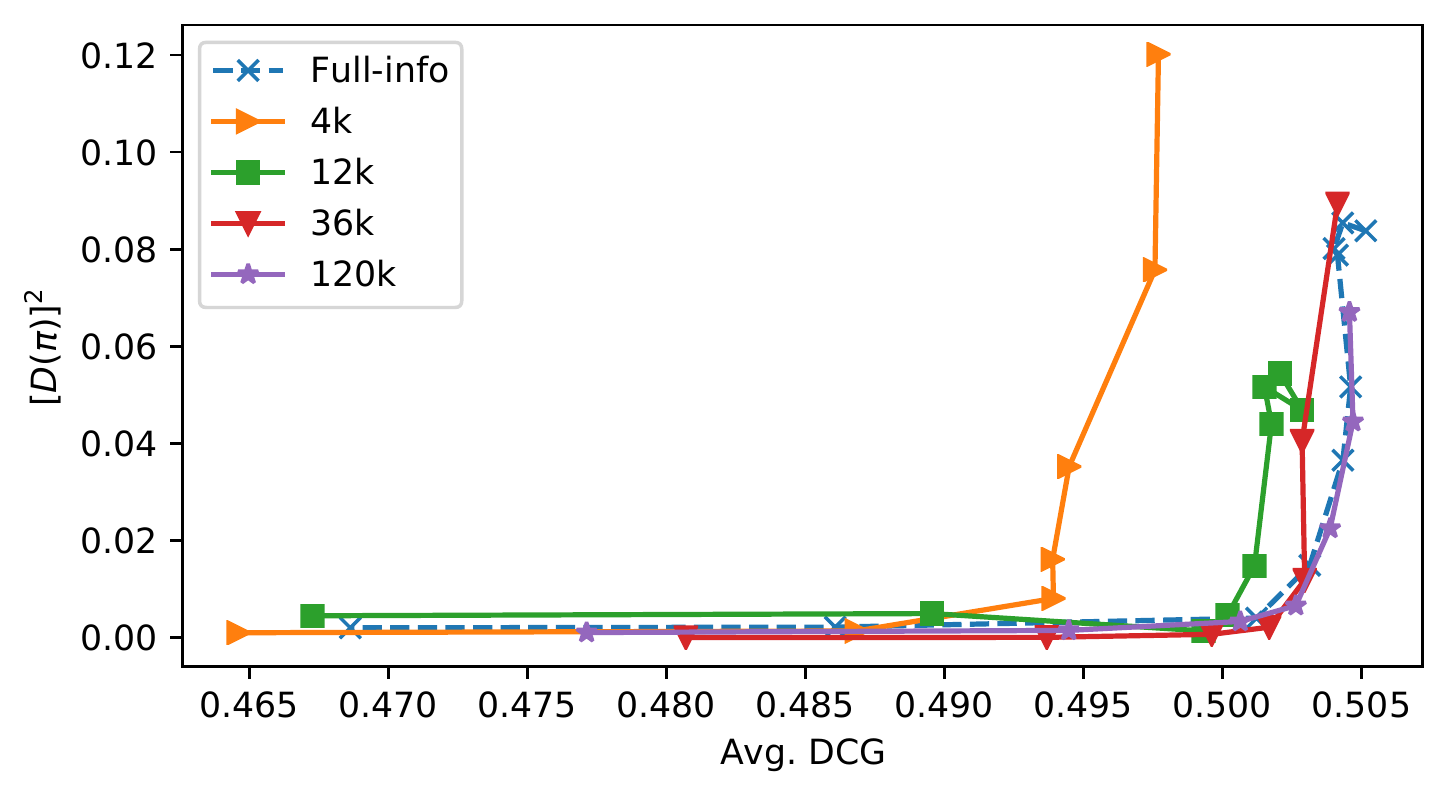}
	\caption{DCG-Disparity curves for \model~ in terms of training clicks on the partial MSLR dataset  ($\eta=1,\epsilon_-=0$).}
	\label{fig:convergence}
\end{figure}
\subsection{Can \model~ converge to the performance of training on the true relevance labels?}

We now explore how ranking utility and fairness of \model~converge as the amount of click data increases. The utility/disparity curves with various amounts of training clicks are shown in \Cref{fig:convergence}. We also show the policy learned by \model~ in the full-info setting, which serves as the skyline that has full knowledge of relevance labels in the training set. With increasing amounts of click data, \model~ approaches the skyline performance of the policy learned on the full-info data. 
The policy learned with 120k clicks is almost identical to that of the full-information policy. This demonstrates that the unbiased estimator of \model~ enables it to converge to the full-information policy given enough click data. 

\subsection{How important is unbiasedness for utility and fairness?}
\begin{figure}[!ht]
	\centering
	\includegraphics[width=0.87\columnwidth]{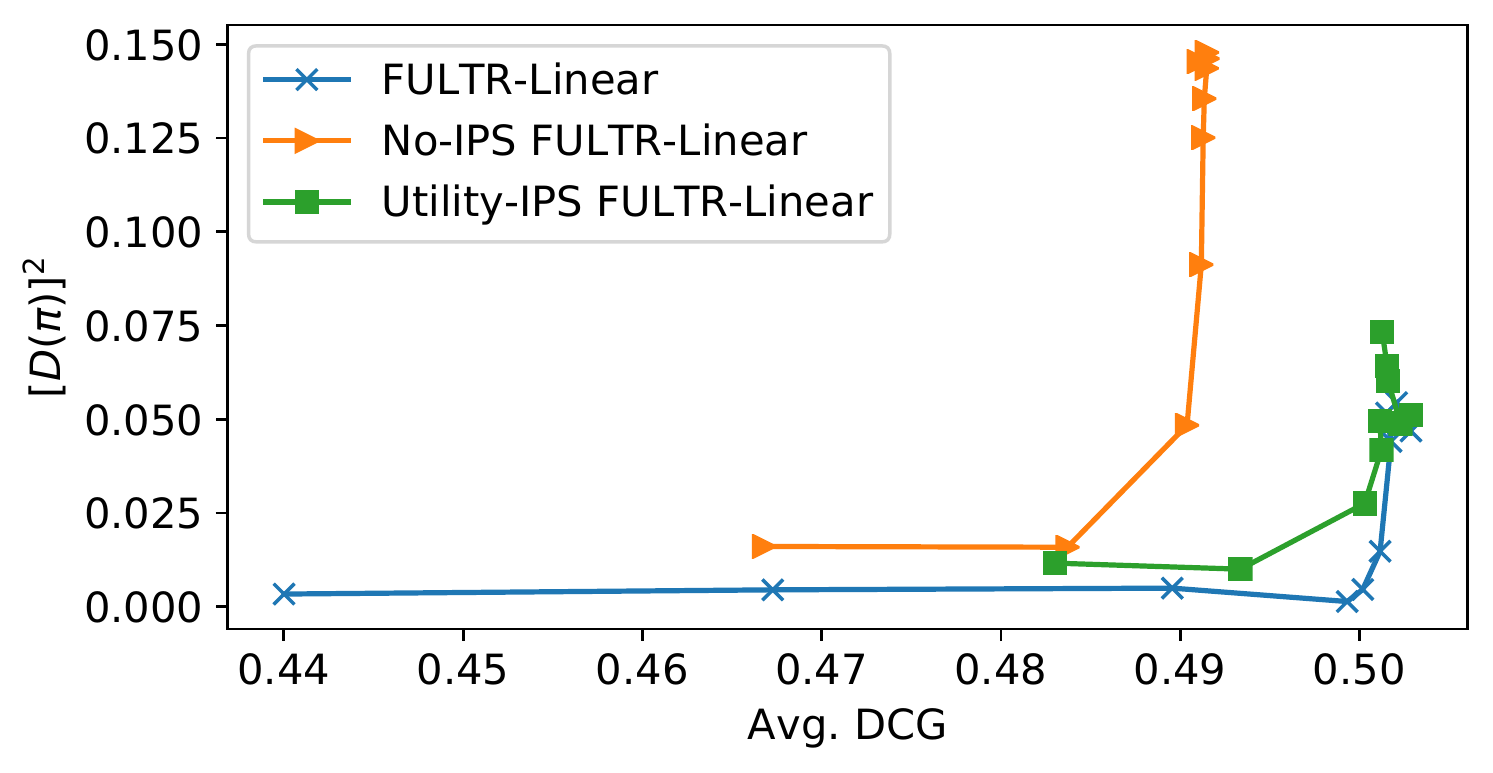}
	\vspace{-0.8em}
	\caption{DCG-Disparity curves for \model~ and two variants with unweighted utility and disparity estimators on the partial MSLR dataset ($N=12K,\eta=1,\epsilon_-=0$).}
	\label{fig:abalation}
\end{figure}
\begin{figure*}[!ht]
	\includegraphics[width=0.9\textwidth]{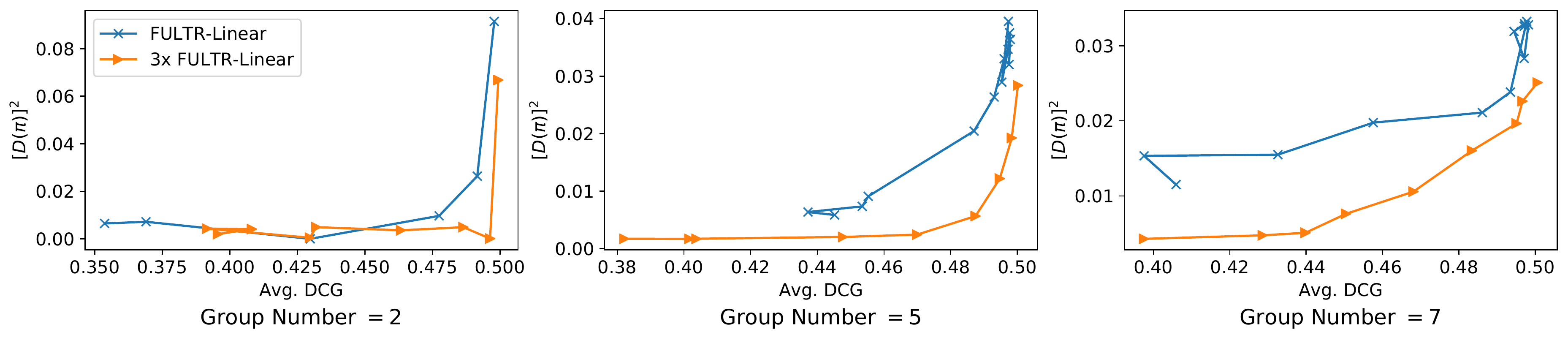}
	\caption{DCG-Disparity curves for different group numbers on the partial MSLR dataset ($N=4K$ and $N=12K$, $\eta=1,\epsilon_-=0$).}
	\label{fig:multiple_group}
\end{figure*}
We conduct an ablation study to understand the effect of IPS-weighting on utility and fairness. In particular, we compare the following variants of \model~ in \Cref{fig:abalation}: 
\begin{itemize}
	\item No-IPS \model-Linear: Both utility and disparity estimators are not IPS-weighted, assuming that implicit feedback reveals full-information relevance labels.
	\item Utility-IPS \model-Linear: The utility estimator is IPS-weighted, but the disparity estimator is unweighted.
\end{itemize}
With biased estimates of both utility and disparity, the \emph{No-IPS} variant of \model~achieves suboptimal utility and cannot reduce disparity to zero. The \emph{Utility-IPS} variant can achieve utility similar to that of \model, but its fairness disparity remains higher even for large values of $\lambda$. 
This implies eliminating position bias through IPS-weighting is essential for both utility and fairness.
\subsection{Can the disparity estimator adjust to click noise?}
\begin{figure}[!ht]
	\centering
	\includegraphics[width=0.87\columnwidth]{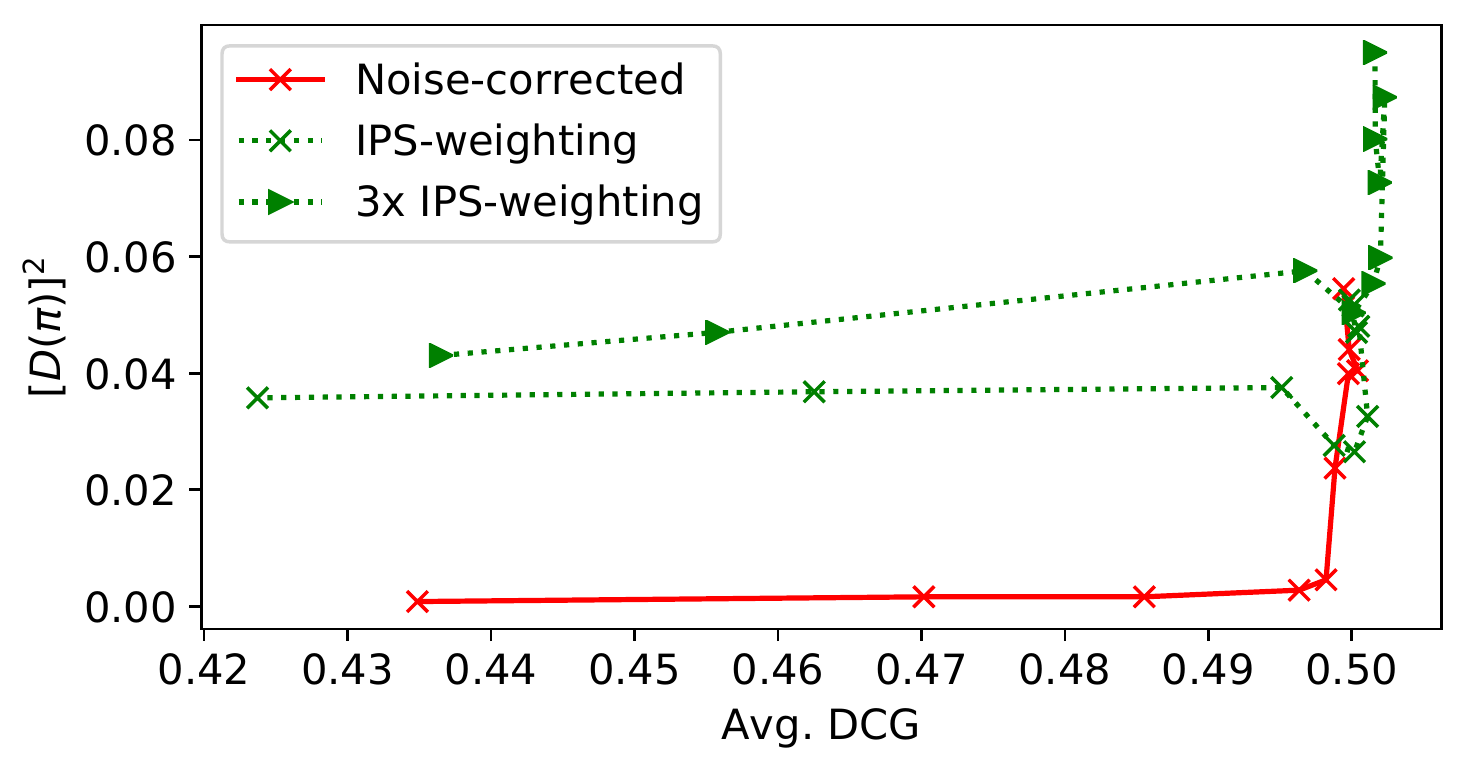}
	\caption{DCG-Disparity curves of noise-corrected estimators against pure IPS-weighting on the partial MSLR dataset with click noise ($N=36K$ and $N=120K$, $\eta=1,\epsilon_-=0.1$).}
	\label{fig:noise}
\end{figure}
Next, we investigate the effectiveness of the noise-corrected disparity estimator in \Cref{eqn:noise-free}. We use partial MSLR with click noise $\epsilon_-=0.1$. Noise-corrected FULTR estimates $\epsilon_-$ via the intervention with 1\% of the training data and applies the noise-corrected estimator in \Cref{eqn:noise-free}. \Cref{fig:noise} shows that the noise-corrected IPS estimator of $D_{ij}(\pi)$ can reduce disparity effectively and achieve zero disparity, while the pure IPS estimator cannot. To verify that this is not due to a lack of data, we increase the amount of training data by a factor of three. However, more data alone does not remedy the problem. This is consistent with the theoretical argument, since more training data can only reduce the variance, while click noise leads to a bias in the disparity estimates.

\subsection{Can \smodel ensure fairness between more than two groups?}
So far, all experiments are conducted with only two groups. To deal with multiple groups, we can modify the constraint in \Cref{eqn:partial_obj} to 
\begin{equation*}
    \sum_{G_i\neq G_j}\Big[\widehat{D}_{ij}(\pi|\queryset)\Big]^2 \leq \delta.
\end{equation*}

To validate the effectiveness of \model~ when dealing with multiple groups, we compare the utility-fairness trade-off for a varying number of groups. To create more than two groups for MSLR, we partition the group attribute into the desired number of intervals with the same numbers of items for each group. 
\Cref{fig:multiple_group} shows the results for 2, 5, and 7 groups. We can observe that with a fixed number of training clicks, the disparity of \model~ increases as the number of groups increases. With 7 groups, a substantial disparity exists even for the largest value of $\lambda$. This is due to a lack of data for each group, as the number of groups is increasing while the amount of training data remains fixed. If we increase the number of training clicks, \model~ can again achieve disparity close to zero even with 7 groups. This demonstrates that \model~ can deal with multiple groups, but that an increase in the number of groups also requires an increase in training data.

